\documentclass[lettersize,journal]{IEEEtran}
\usepackage{amsmath,amssymb,amsfonts}

\usepackage{graphicx}
\usepackage{textcomp}
\usepackage{array}
\usepackage[caption=false,font=normalsize,labelfont=sf,textfont=sf]{subfig}

\usepackage{stfloats}
\usepackage{url}
\usepackage{verbatim}

\usepackage{epstopdf}
\usepackage{balance}

\usepackage{CJK}
\usepackage[top=2cm, bottom=2cm, left=2cm, right=2cm]{geometry}
\usepackage{algorithm}
\usepackage{algorithmicx}
\usepackage{algpseudocode}
\usepackage{amsmath}

\usepackage{multirow}
\usepackage{booktabs}

\floatname{algorithm}{Algorithm}

\usepackage{amssymb}

\usepackage{color}
\usepackage{threeparttable}

\usepackage{indentfirst}

\usepackage{caption}
\usepackage[utf8]{inputenc}
\usepackage[T1]{fontenc}
\usepackage{enumerate}
\newsavebox{\tablebox}
\usepackage{graphicx}

\usepackage{amssymb}

\usepackage{threeparttable}

\usepackage{multirow}

\usepackage{stfloats}
\usepackage{float}

\usepackage{soul}
\usepackage[colorlinks=false,linkcolor=cyan,filecolor=blue,urlcolor=red,citecolor=green]{hyperref}

\def\BibTeX{{\rm B\kern-.05em{\sc i\kern-.025em b}\kern-.08em
    T\kern-.1667em\lower.7ex\hbox{E}\kern-.125emX}}
\begin{document}
\title{Partial Connection Based on Channel Attention for Differentiable Neural Architecture Search}

\author{Yu Xue, \IEEEmembership{Member, IEEE}, Jiafeng Qin	
\thanks{This work was partially supported by the National Natural Science Foundation of China
		(61876089, 61876185, and 61902281), the Natural Science Foundation of Jiangsu Province (BK20141005).}		
\thanks{Yu Xue (corresponding author) and Jiafeng Qin are with the School of Computer Science, Nanjing University of Information Science and Technology, Jiangsu, China. E-mails: xueyu@nuist.edu.cn; qinjiafeng@nuist.edu.cn.}
}

\maketitle

\begin{abstract}
Differentiable neural architecture search (DARTS), as a gradient-guided search method, greatly reduces the cost of computation and speeds up the search. In DARTS, the architecture parameters are introduced to the candidate operations, but the parameters of some weight-equipped operations may not be trained well in the initial stage, which causes unfair competition between candidate operations. The weight-free operations appear in large numbers which results in the phenomenon of performance crash. Besides, a lot of memory will be occupied during training supernet which causes the memory utilization to be low. In this paper, a partial channel connection based on channel attention for differentiable neural architecture search (ADARTS) is proposed. Some channels with higher weights are selected through the attention mechanism and sent into the operation space while the other channels are directly contacted with the processed channels. Selecting a few channels with higher attention weights can better transmit important feature information into the search space and greatly improve search efficiency and memory utilization. The instability of network structure caused by random selection can also be avoided. The experimental results show that ADARTS achieved 2.46\% and 17.06\% classification error rates on CIFAR-10 and CIFAR-100, respectively. ADARTS can effectively solve the problem that too many skip connections appear in the search process and obtain network structures with better performance.
\end{abstract}


\section{Introduction}
\label{sec:introduction}
\IEEEPARstart{D}{eep} neural network \cite{9229504, slowik2010application} is an important research topic in deep learning \cite{deng2021deep}. Compared with shallow neural network, its multi-layer network structure can extract richer and more complex feature information to obtain higher performance. Therefore, it has made remarkable progress in semantic recognition \cite{8205956}, image recognition \cite{ 8972242},  and data forecasting \cite{ 9376913, 9145791}. The performance of deep neural network mainly depends on network parameters and structure. The vital task is to design appropriate strategies to optimize the structure and parameters of the neural network to improve its final performance. At present, gradient methods are mainly used to optimize network parameters, such as SGD optimizer \cite{yu2019linear} and ADAM optimizer \cite{DBLP:journals/corr/KingmaB14}. 
 However, unlike parameter optimization, how to build a better performance network structures requires a lot of expert experience. For example, some of the better neural networks, such as VGG \cite{simonyan2014very} and ResNet \cite{he2016deep}, were designed by experts. But, all of them are designed manually by using the trial and error manner. A lot of time and resources are taken to design these networks, and different network structures need to be built for different problems and datasets. Neural architecture search (NAS) \cite{9684972} as an effective method can automatically search network structures with higher performance.

As a part of automated machine learning (AutoML), the NAS builds different neural network structures by constructing a large search space and search strategies. In NAS, the search space, the appropriate search method, and the evaluation method are three main tasks.

\begin{figure}[htbp]
	\centering

	\includegraphics[width=8cm]{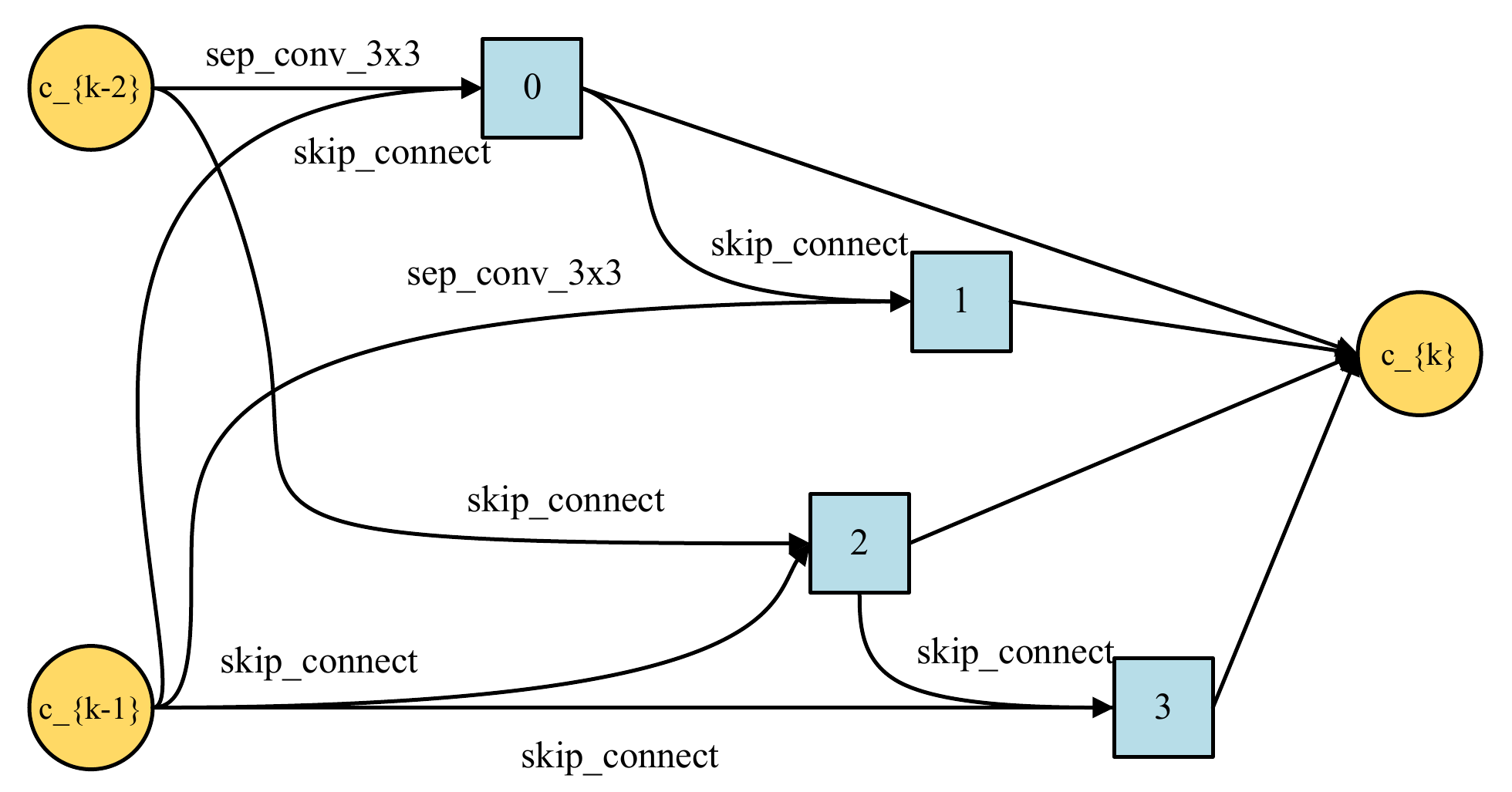}
	\caption{Searched normal cell on Cifar-10 which has too many skip connections.}
	\label{f01}
\end{figure}

In general, there are many methods that can be used for architectural search, such as reinforcement learning (RL) \cite{zoph2018learning}, evolutionary computing (EC) \cite{xie2017genetic} and gradient-based. 
Reinforcement learning regards the search of neural network structure as an agent's action. The network is constructed through different behaviors and rewards which are based on the evaluation of the network performance on the test set. Representing and optimization of the agent policy are the two keys to using reinforcement learning to search network architecture. Zoph~\emph{et~al.} \cite{zoph2017neural} used recurrent neural network (RNN) strategies to sequentially sample strings and then encode the neural architecture. 
Baker~\emph{et~al.} \cite{baker2016designing} used a q-learning training strategy, which in turn selects the type of layer and corresponding hyperparameters. However, the method based on reinforcement learning consumes extremely computing resources. For example, Zoph \cite{zoph2017neural} used 800 GPUs to complete the search process for three to four weeks. Therefore, as another method to replace reinforcement learning, evolutionary computing can reduce computational consumption and find a better solution compared with reinforcement learning. Xie~\emph{et~al.} \cite{xie2017genetic} represented network structure with a fixed length binary code, and used genetic algorithm to initialize individuals and explore network space through selection, crossover and mutation. It only used 17 GPUs to train for one day on the same dataset which is much faster than the reinforcement learning method. The difference of evolutionary computation methods mainly exists in how to choose the initial population, update the population, and generate offspring. Real~\emph{et~al.} \cite{real2017large} used the tournament method to select the parent and deleted the worst individual from the population, and the new offspring inherited all the parameters of the parent. Although such inheritance is not strict with inheritance performance, it can also speed up the learning process compared with random initialization. Elsken~\emph{et~al.} \cite{elsken2018efficient} adopted the sample parent from the multi-objective Pareto frontier to generate better offspring. In terms of initialization methods, these algorithms often encoded convolution, pooling and other operations directly \cite{sun2019evolving}.
In addition, the basic modules used in ResNet and DenseNet networks can well deal with gradient disappearance,  
Sun~\emph{et~al.} \cite{sun2019completely} used the genetic algorithm to search ResNet and DenseNet blocks to get constructed network with high performance. Yang~\emph{et~al.} \cite{yang2020cars} used the cell which is a continuous search space as a basic module to speed up the search. However, the method based on evolutionary computation still has considerable computational overhead, and it takes a lot of time to evaluate multiple network structures. To further reduce search cost, 
differentiable architectural search (DARTS) \cite{liu2018darts} used weight sharing and 
optimized the process of supernet training and subnet search so that it can search for the neural network architecture quickly. However, DARTS has the phenomenon that the weight-free operations increase in the later stage of the search, such as skip connection and max pool. A large number of skip connections appear in a cell in Fig. \ref{f01}. This type of cell cannot extract image features well , resulting in network performance collapse.

To search for the network architecture with high performance and further improve search efficiency, this paper proposes a partial channel connection based on attention mechanism for differentiable neural architecture search (ADARTS). The channel attention mechanism is used to select important channels and reinforce feature information. In addition, partial channel connections reduce unfair competition between operations and reduce memory occupation, making the search process faster and more stable. We summarize our contributions as follows:

	1. The attention mechanism is employed to extract the importance of the channels of the input data in the searching process. Then, the obtained attention weights are multiplied by the original input data to generate new input so that the key features in the input data can be identified and the neural network can use more important information.

	2. The channel selection is used to send the channels with higher attention weights into operation space, and the other channels are directly contacted with the output of operation space to improve the search efficiency and stability. Furthermore, it can weaken the unfair competition between candidate operations caused by the parameters of weight-equipped operations which are not trained well in the initial stage.

	3. The proposed method searches 0.2 GPU days on CIFAR-10 dataset, and the searched structure achieves 2.46\% classification error and 2.9M parameters. It also achieves 17.06\% classification error when transferred to CIFAR-100 dataset for evaluation, which is better than other comparative NAS methods.

The remainder of this article is organized as follows. In Section \uppercase\expandafter{\romannumeral2}, we introduce the basic ideas and methods of DARTS. In section \uppercase\expandafter{\romannumeral3}, we describe the proposed algorithm that uses the channel attention strategy to select some channels for connection. Section \uppercase\expandafter{\romannumeral4} introduces the experimental plan, datasets, parameters, and accuracy in image classification. In addition, we prove its effectiveness through comparison with other NAS algorithms and ablation studies. Finally,  the conclusion is expressed in section \uppercase\expandafter{\romannumeral5}. 

\begin{figure*}[htbp]
	\centering

	\includegraphics[width=15cm]{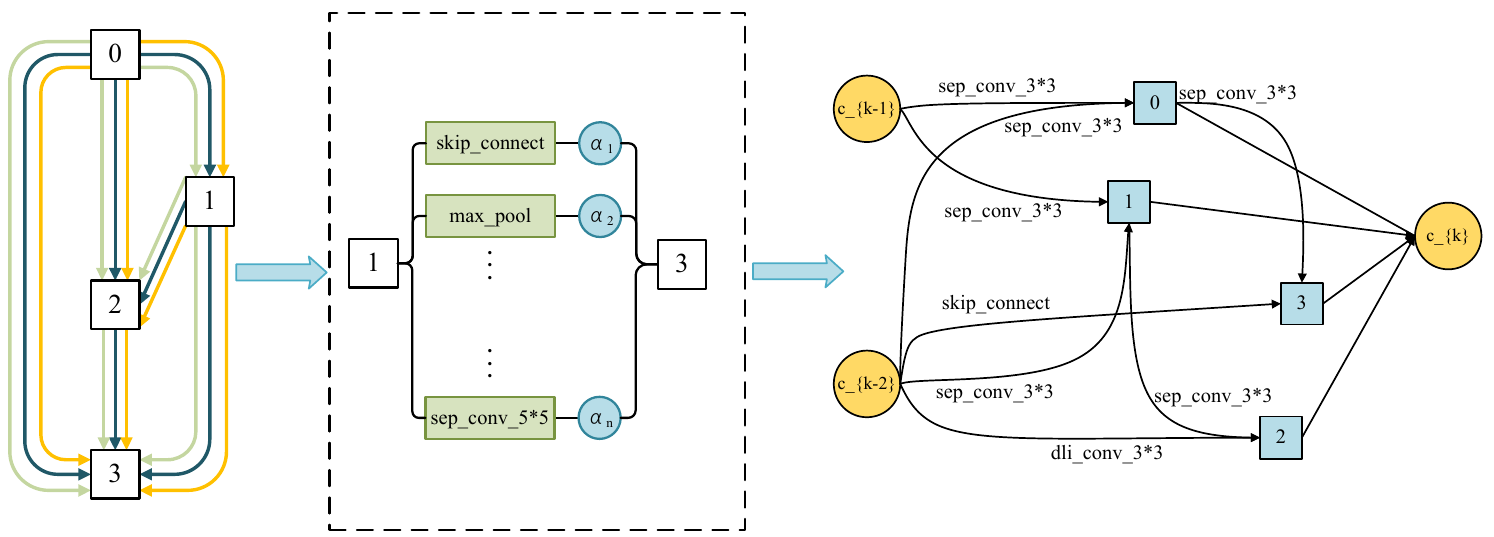}
	\caption{The process of differentiable neural architecture search, the left illustration is the search space of the cell, the middle illustration is the operation space between node $i$ and node $j$, the right illustration is the selected structure by selecting the operation with max weigh $\alpha$ between node $i$ and node $j$. }
	\label{f02}
\end{figure*}
\section{Related Works}
Differentiable neural architecture search makes the search space continuous and uses gradients to alternately optimize network parameters and network structure weights, which greatly reduces the cost of calculation and improves the search speed. Fig. \ref{f02} describes the basic process of DARTS. First, the search space is composed of nodes, and there is a candidate operation space between the nodes. DARTS needs to select the connection mode between two nodes. The operation with the highest weight in the candidate space is selected as the connection operation. After all connected directed edges and corresponding operations are selected, the final network structure is determined. However, unfair competition in the operation space is caused because the weight-equipped operations in the initial training stage are not trained well, which leads to the network collapse. There are a large number of skip connections in the network and the performance of the network drops sharply. 
FairDARTS \cite{chu2020fair} used sigmoid functions instead of softmax functions to calculate weights to avoid unfair competition between operations. PDARTS \cite{chen2019progressive} used dropout to randomly cut skip connections to mask some skip connections during network searching and reduced the proportion of dropout with the gradual training of other operating parameters. In addition, a progressive search method was introduced, and the number of cells was gradually increased to make it more similar to the final trained network. However, a larger memory occupation was generated when the cells were increased. PDARTS continuously reduced the candidate operations in the training process, but the final structure could be restricted. To increase the memory utilization of DARTS in searching, GDAS \cite{dong2019searching} only sampled a part of the operation each time for optimization. PCDARTS \cite{xu2019pc} adopted the method of partial channel connection to reduce the operation in the operation space, but the random selection of the channel leaded to the instability of the network structure. The proposed method in this paper selects more important channels by channel attention weight to update the operation weight, making the search more stable and faster.
\subsection{Search Space}

In DARTS, some cells are stacked to form the network structure. Cells are divided into reduction cells and normal cells. The same type of cell has the same internal connection mode and share weight. A cell is a directed graph containing \emph{k} nodes x$_{i}$=\{x$_{1}$, x$_{2}$, ... ,x$_{k}$\}, two input nodes, one output node and \emph{k-3} intermediate nodes. In addition, the \emph{j$^{th}$} intermediate node is connected to the all predecessor nodes respectively and the input of the middle node is obtained through its predecessor nodes, as shown in formula (1):
\begin{equation}
	x_{j}=\sum_{i<j}o^{(i,j)}(x_i)
\end{equation}
where $x_j$ is the $j^{th}$ node, $x_i$ represents the $i^{th}$ node, $o(i,j)$ represents the candidate operation between node $j$ and node $i$.

Each directed edge $(i,j)$ has corresponding operation $o(i,j)$ and weight $\alpha_{o^{(i,j)}}^k$ which is calculated by softmax, as shown in Formula (2). The output of the candidate operation is weighed with the corresponding weight $\alpha_{o^{(i,j)}}^k$. The total input $f(x^j)$ is expressed as formula (3):
\begin{equation}
	\alpha_{o^{(i,j)}}^k=\frac{exp(\alpha_{o^{(i,j)}}^{k})}{\sum_{k'\in o}exp(\alpha_{o^{(i,j)}}^{k'})}
\end{equation}
\begin{equation}
	f(x_j)=\sum_{k\in o} \alpha_{o^{(i,j)}}^ko^{(i,j)}(x_i)
\end{equation}
where $ o $ is the candidate operation space, including $ 'none' $, $ 'maxpool\_3\times3' $, $ 'avgpool\_3\times3' $, $'skip\_connect'$, $'sepconv\_3\times3'$, $'dilconv\_3\times3'$, $'sepconv\_5\times5'$, $'dilconv\_5\times5'$, $ o^{(i,j)} $ is the candidate operation, $ \alpha_{o^{(i,j)}}^k $ is the weight of candidate operation on each directed edge $ (i,j) $.
\subsection{Search Method}

There are two groups of parameters in DARTS. In addition to network weight $\omega$ was used to train the network. Besides, architecture parameter $\alpha$ is added to transform the discrete space into a continuous search space and the parameters are trained by gradient descent. The training loss \emph{$\mathcal{L}$$_{train}$} is used to optimize the network weight parameter $\omega$, and the valid loss \emph{$\mathcal{L}$$_{val}$} is used to optimize the architecture parameter $\alpha$. The two optimization methods are carried out as formula(4-5) and finally the subnet with better performance is searched:
\begin{flalign}
	\hspace{0.36\linewidth}&\min_{\alpha} \ \      \mathcal{L}_{val}(\omega^*(\alpha),\alpha)&
\end{flalign}
\begin{equation}
	s.t.\ \  \omega^*(\alpha) = argmin_\omega \ \  \mathcal{L}_{train}(\omega,\alpha)
\end{equation}
where $\mathcal{L}_{val}$ is valid loss, $\omega$ is network weight, $\alpha$ is network architecture weight, $\mathcal{L}_{train}$ is training loss.

After the $\omega$ and $\alpha$ parameters are updated in each generation, for each intermediate node $x_j$, two connecting edges are selected according to the largest $\alpha_{o^{(i,j)}}^k$ on the directed edge $(i,j)$ of the node, and the corresponding operation $o_{select}^{(i,j)}$  will be selected. After the connection mode and operation of all intermediate nodes are determined, other directed edges and candidate operations will be removed. The method of selecting operation is as follows:
\begin{equation}
	o_{select}^{(i,j)}=argmax_{k\in o} \ \ \alpha_{o^{(i,j)}}^k
\end{equation}
where $o_{select}^{(i,j)}$ is the selected operation between node $i$ and node $j$, $\alpha_{o^{(i,j)}}^k$ is the corresponding weight of operation.
\section{METHODOLOGY}
In the search process of DARTS, each operation and its corresponding output need to be stored in the node, which occupies a large amount of memory. Therefore, the researchers have to set a small batchsize, which limits the memory utilization. The proposed partial channel connection method can greatly improve the operation efficiency and reduce the unfair competition between candidate operations. However, the random selection of channels also leads to instability in the search process. Therefore, this paper proposes to use the attention mechanism to select more meaningful channels and strengthen the features of some channels to make the searched network structure more stable. In the following subsection, we firstly introduce the framework of our algorithm, and then elaborate the important steps of attention mechanism, channel selection and partial channel connection in turn.
\subsection{Overall Framework}

 The framework of ADARTS is simply described as follows: firstly, the attention mechanism is added before each operation space to produce inputs, i.e., the features of all channels are extracted through global pooling and used as the inputs of a multi-layer perceptron (MLP). Then channel attention weights $F_c$ are obtained through the MLP. After that, the obtained weight $F_c$ is multiplied by the original input to get the new input. Finally, the top $1/K$ new channels with the larger weights are marked as 1 and the rest marked as 0, and the channels marked with 1 are sent into the operation space of ADARTS for the following calculation, while the other channels are directly contacted with the output.

\subsection{Attention Mechanism}

\begin{figure*}[htbp]
	\centering
	\includegraphics[width=0.8\linewidth]{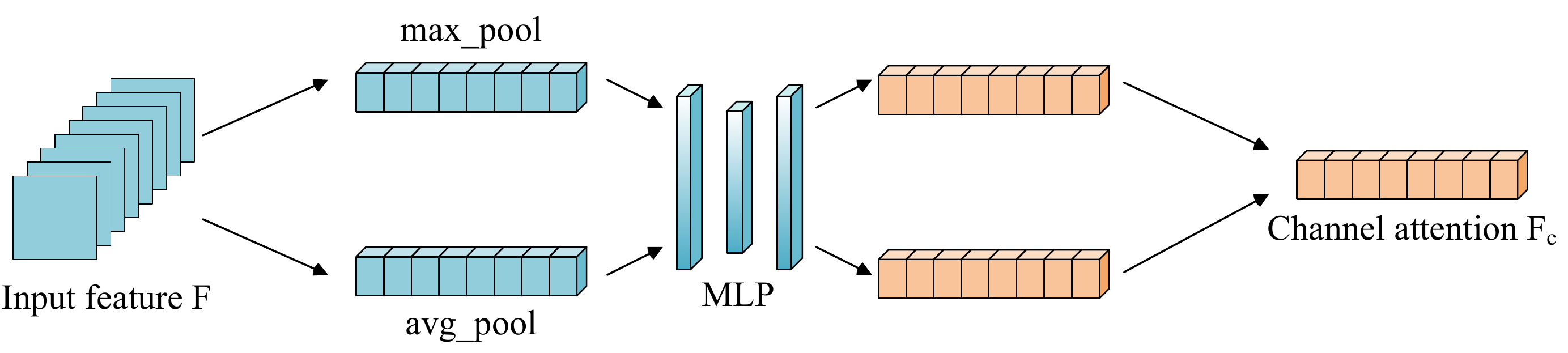}
	\caption{The process of calculating channel attention. The input feature F will be processed with maxpool and avgpool respectively, then go through MLP. The corresponding output of MLP will be added to generate channel attention weight $F_c$.   }
	\label{f03}
\end{figure*}

Usually, many channels are generated by convolution operations in deep neural networks, but some feature information from some channels has little effect. Channel attention mechanism focuses on distinguishing the importance of channels \cite{woo2018cbam}. In the channel attention mechanism, it is necessary to compress channels to extract feature information. Therefore, max pooling and average pooling are used to obtain spatial features. The pseudo codes of attention mechanism are given in Algorithm 1. For an input image data $F$ with height of $H$, width of $W$, and channels of $C$ in the network, max pooling and average pooling are used for each channel respectively to obtain the feature data $f_{ap}$ and $f_{mp}$ with the size of 1$\times$1$\times$$C$. Then the two feature data are input into a shared MLP. The neurons of the hidden layer are set to \textit{C/K} to reduce computational complexity. Through MLP, $f'_{ap}$ and $f'_{mp}$ obtained are added to generate channel attention weight $F_c$. The calculation process of the channel attention weights is shown in Fig. \ref{f03}. The map function of channel attention mechanism is given as the formula (\ref{l7}) as follows:
\begin{equation}
	\begin{aligned}
		F_c &=\sigma(MLP(avgpool(F))+MLP(maxpool(F)))\\
		&= \sigma(MLP(f_{ap})+MLP(f_{mp}))\\
		&= \sigma(\omega_2(\omega_1(f_{ap})))+\omega_2(\omega_1(f_{mp})))\\
		&= f'_{ap}+f'_{mp}
		\label{l7}
	\end{aligned}
\end{equation}
where $F$ is input image data, $f_{ap}$ is average pooled data $F$, $f_{mp}$ is max pooled data $F$, MLP is the multi-layer perceptron, $\omega_1$ are weights between input layer and hidden layer, $\omega_2$ are weights between hidden layer and output layer, $f'_{ap}$ is the result of $f_{ap}$ processed by MLP, $f'_{mp}$ is the result of $f_{mp}$ processed by MLP, ${F_c}$ is the channel attention weight, and $\sigma$ is a sigmoid function which is $1 / (1+e^{-x})$.

\begin{figure}[htbp]
	\centering
	\includegraphics[width=7cm]{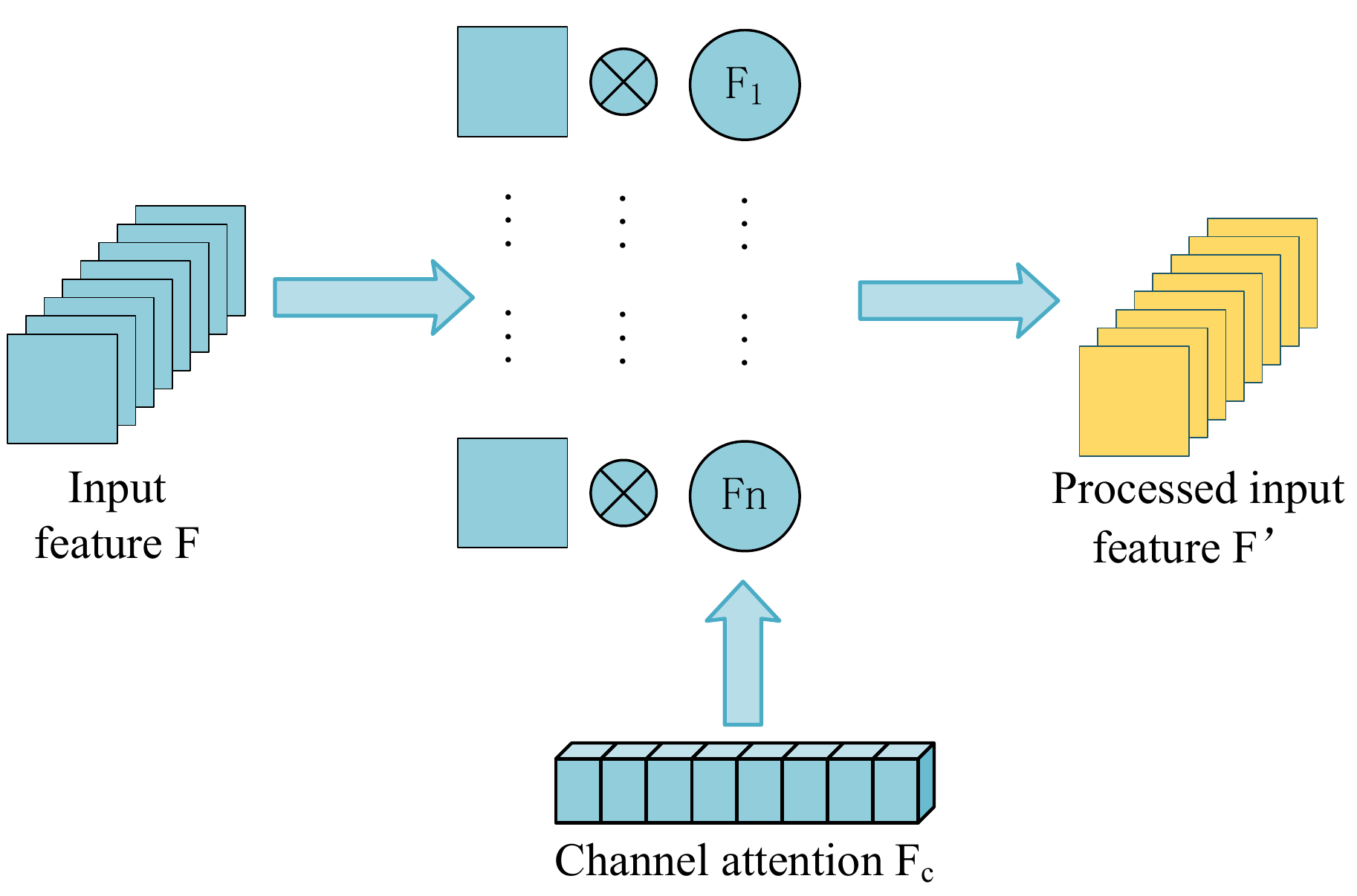}
	\caption{The process of enhancing feature of input data F by channel attention $F_c$.}
	\label{f04}
\end{figure}

Then, new feature maps are generated by multiplying the channel attention weight and the original feature maps, so the importance of input features can be identified. In detail, Fig. \ref{f04} describes the calculation process that the data $F$ with the original input size of $H$$\times$$W$$\times$$C$ is multiplied with $F_c$ of $1$$\times$$1$$\times$$C$ in channel order to obtain $F'$. The calculation method is expressed as formula (\ref{l8}):
\begin{equation}
	F'=F_c*F
	\label{l8}
\end{equation}
where $F_c$ is channel attention weight, $F$ is input feature data, $F'$ is feature data multiplied by channel attention weight.
\begin{algorithm}
	\caption{Attention Mechanism}
	\begin{algorithmic}[1] 
		\Require Feature data  $F$
		\Ensure Result data $F'$ of channel attention 
		\Function {Attention}{$F$}
		\State $f_{ap} \gets$ Global average pooling of $H$$\times$$W$$\times$$C$ feature  \Statex \qquad \qquad \ \ data $F$
		\State $f_{mp} \gets$ Global max pooling of $H$$\times$$W$$\times$$C$ feature  \Statex \qquad \qquad \ \  data $F$
		\State $f'_{ap} \gets$  $1$$\times$$1$$\times$$C$ feature data $f_{ap}$ is sent into MLP \Statex \qquad \qquad \ \ for calculation
		\State $f'_{mp} \gets$  $1$$\times$$1$$\times$$C$ feature data $f_{mp}$ is sent into MLP \Statex \qquad \qquad \ \ for calculation
		\State $F_c \gets$  Add the $f'_{ap}$ and  $f'_{mp}$ to get channel attention \Statex \qquad \qquad \ \ weight
		\State $F' \gets$ Multiply $F_c$ with the original feature data $F$ \Statex \qquad \qquad \ \ according to the channel order
		\State \Return{$F'$}
		\EndFunction
	
	\end{algorithmic}
\end{algorithm}
\subsection{Channel Selection}
\begin{figure}[htbp]
	\centering
	\includegraphics[width=5.1cm]{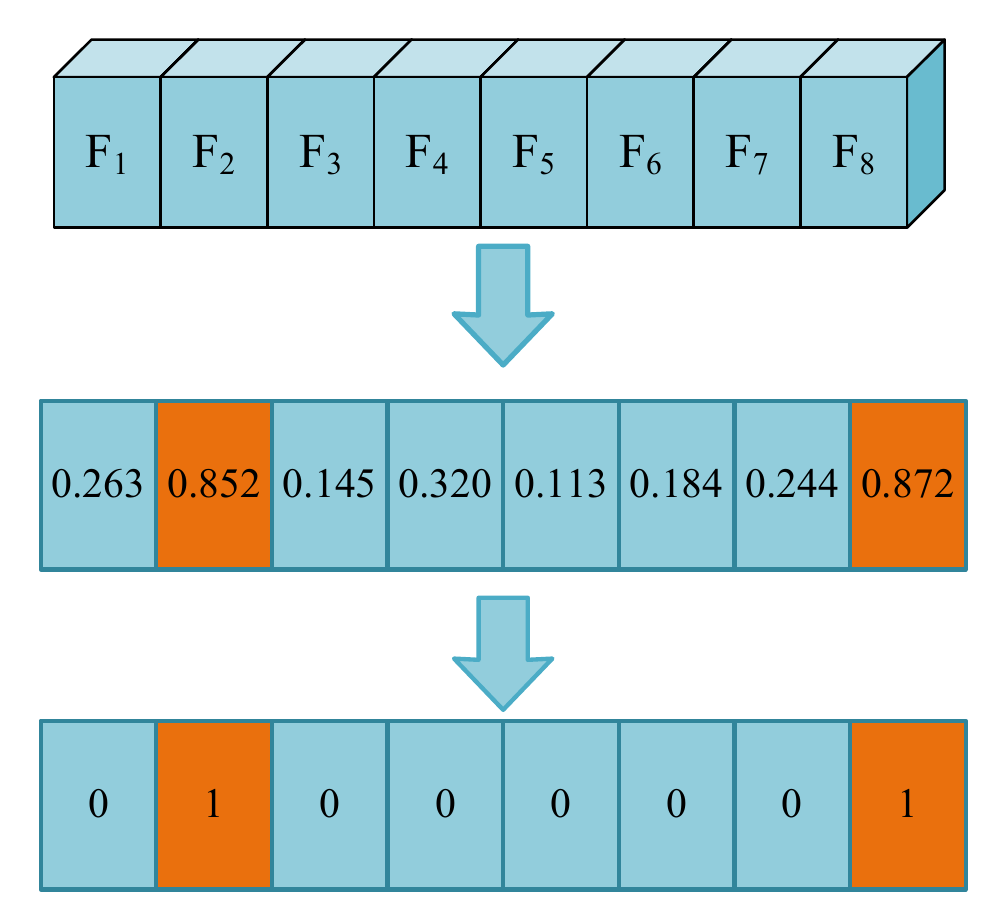}
	\caption{The process of selecting channels according to channel attention weight. The top $1/K$ channels are selected and marked as 1 and the rest as 0. }
	\label{f05}
\end{figure}

The weights of channel attention can reflect the importance of feature channels. The channels with higher attention weight possess more important information. To better carry out partial channel connections, the more important channels need to be selected to make the search more stable and accurate. In this paper, channel mask $M^{(i,j)}$ which is computed according to formula (9) is used to represent the selected feature channels and masked feature channels in the operation space $o^{(i,j)}$. The pseudo codes of feature channel selection are shown in Algorithm 2. In the process, the top $1/K$ channels will be selected according to the weight of channel attention and the rest are masked channels. The selected channels will be assigned a value of 1 and transferred to the operation space for calculation. The masked channels will be assigned a value of 0 and will directly skip the operation space and be concatenated with the output. This feature channel process is shown in Fig. \ref{f05}.
\begin{equation}
	M^{(i,j)}=\left\{
	\begin{array}{rcl}
		1       &      & {F_k\in 1/K \  channels}\\
		
		0       &      & {F_k\notin  1/K \ channels}
	\end{array} \right. 
\end{equation}
where $M^{(i,j)}$ is the channel mask between $i^{th}$ node and $j^{th}$ node, $F_k$ is $k^{th}$ channel attention weight.
\begin{algorithm}
	\caption{Channel selection}
	\begin{algorithmic}[1] 
		\Require Channel attention $F_c$, selected channel  proportion \Statex \quad \ \   $1/K$
		\Ensure Channel mask $M^{(i,j)}$
		\Function {Selection}{$F_c,K$}
		\State $k \gets 0$
		\State $F_{max} \gets top\  1/K\  weight\  in\  F_c$
		
		\While {$k < c$}
		\If{$F_k \in F_{max}$}
		\State $M_k^{(i,j)} \gets 1$
		\Else
		\State $M_k^{(i,j)} \gets 0$  
		\EndIf
		\State $k \gets k+1$
		
		\EndWhile

		\State \Return{$M^{(i,j)}$}
		\EndFunction
		
	\end{algorithmic}
\end{algorithm}

\subsection{Partial Channel Connection}

\begin{figure*}[htbp]
	\centering
	\includegraphics[width=14cm]{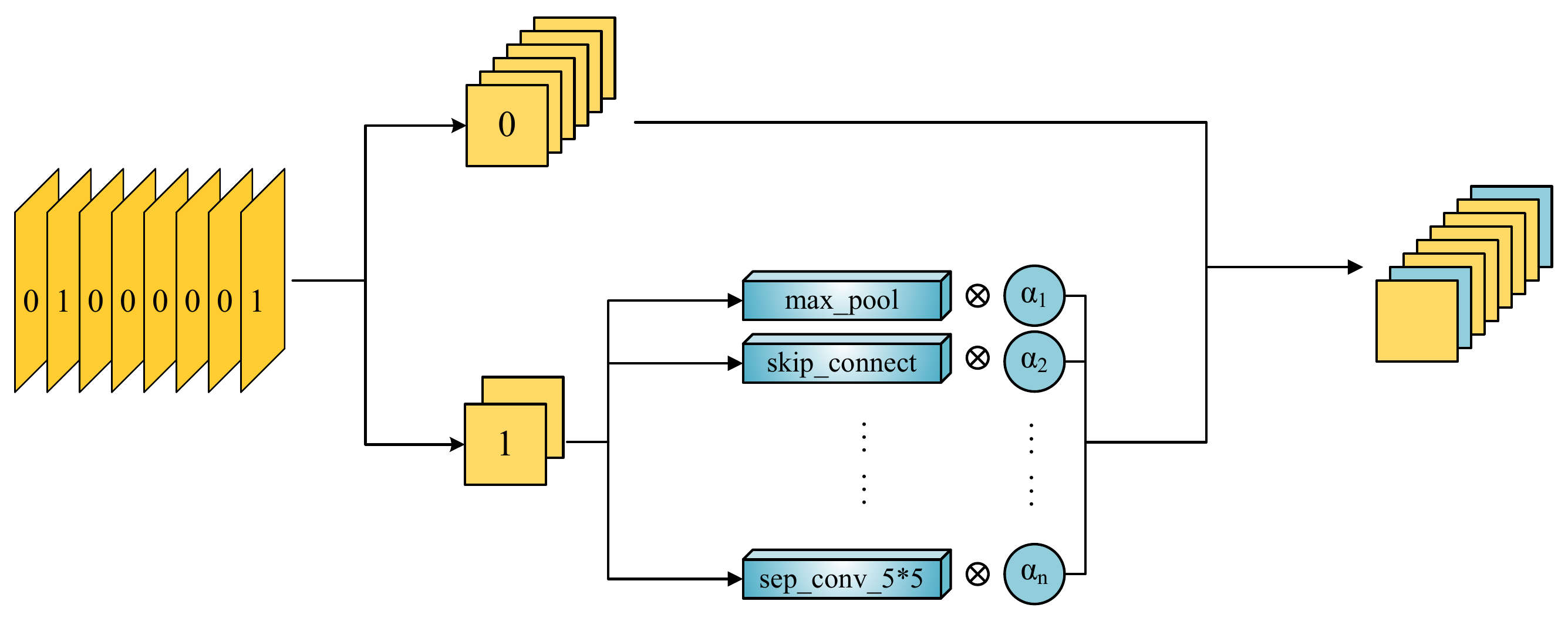}
	\caption{The selected channels are marked as 1 and the masked channels are marked as 0 according to channel attention weight. The selected channels will go through operation space, and a calculation result of each channel is multiplied by the weight $\alpha$. Finally, the output channels are contacted with the remaining channels.    }
	\label{f06}
\end{figure*}

To solve the problem that too small batchsize will cause the instability of parameters and structures during the search process, this paper proposes partial channel connections to reduce the memory usage in operation space at runtime. On the same batchsize, if the number of partially connected channels is $1/K$, the running memory usage will be reduced to the original $1/K$. In this way, the batchsize of input data can be greatly increased, which can speed up the operation and make the network search more stable. Besides, DARTS tends to choose weight-free operations. This is because the weight-free operations provide more precise information than the weight-equipped operations when the neural network is not well trained. For example, the skip connection is to transfer data directly to the next node. Because the weight training needs a lot of iterations, weight-free operations will accumulate a lot of advantages before training completely. In addition, there is a competitive relationship between operations, so even in the late training period, the weight-free operations have more advantages, which results in the phenomenon of network structure collapse. The pseudo codes of the partial channel connection are provided in Algorithm 3. In this algorithm, by using partial channel connections, only the selected channels enter the operation space for training, while the rest of the channels are not processed and then contacted with the outputs of the selected channels. The process is described in Fig. \ref{f06}. The calculation of output of operation space $o^{(i,j)}$ between node $j$ and node $i$ is expressed by formula (10). $M^{(i,j)}$ represents the selected and masked channels. Even weight-equipped operations are not trained well, the loss of using partial channels is smaller than the method that all channels enter the operation space. It makes the advantage of weight-free operations less obvious. However, because only the partial channels are sent into the operation space, random selection of connection channels will also cause instability of network architecture and parameters. Therefore, channel attention weight $F_c=\{F_1, F_2... F_n\}$ in Section \uppercase\expandafter{\romannumeral3} (B) is used when selecting channels. The selection of $1/K$ channels with larger weights as the input channel can not only strengthen the data feature and speed up the search efficiency, but also make the weight parameters more stable during training to strengthen the searched network structure and avoid the phenomenon of structural collapse in the later search.

\begin{equation}
	f(x_j)=\sum_{k\in o} \alpha_{o^{(i,j)}}^ko^{(i,j)}(x_i*M^{(i,j)})+(1-M^{(i,j)})*x_i
\end{equation}
where $x_i*M^{(i,j)}$ are selected channels, $(1-M^{(i,j)})*x_i$ are masked channels, $f(x_j)$ is the input of node $j$, $ \alpha_{o^{(i,j)}}^k $ is the weight of candidate operation on each directed edge $ (i,j) $, $o^{(i,j)}$ is the candidate operation between node $j$ and node $i$. 
\begin{algorithm}
	\label{a03}
	\caption{Partial Channel Connection}
	\begin{algorithmic}[1] 
		\Require Feature data through channel attention $F'$, channel \Statex \quad \ \  mask $M^{(i,j)}$
		\Ensure  Hybrid computation $f(x_j)$ of node $j$ 
		\Function {Partialconnection}{$F', M^{(i,j)}$}
		\State $k \gets 0$
		\While {$k < c$}
		\If {$M_k^{(i,j)}==1$}
		\State $F'_k$ is sent to operation space to perform \Statex \qquad \qquad \ \    hybrid operations.
		\Else 
		\State $F'_k$ skips the operation space
		
		\EndIf
		\State $k$++
		\EndWhile
		\State The feature output through the operation space \Statex \quad \ \  and the unprocessed feature are contacted to produce   \Statex \quad \ \ $f(x_j)$ according  to the original channel order.
		\State \Return{$f(x_j)$}
		\EndFunction
		
	\end{algorithmic}
\end{algorithm}

\section{EXPERIMENTS ON CLASSIFICATION TASKS}
We applied ADARTS to image classification to test its performance by using common image datasets (CIFAR-10/ CIFAR-100). The network architecture found on CIFAR-10 is transferred to CIFAR-100 for evaluation. In addition, we make ablation experiments to verify the classification accuracy and stability of ADARTS.

\subsection{Dataset}
CIFAR-10 and CIFAR-100 have 60,000 color images with 32$\times$32 resolution. There are 10 classes with 6000 images each in CIFAR-10 and 100 classes with 600 images each in CIFAR-100. Each dataset is divided into two parts, i.e., training set and test set. The training set is used to search for network structure, and the test set is used to verify the performance of the network.  

\begin{figure}[htbp]
	\centering	
	\subfloat[ \ ]{
		\centering
		\begin{minipage}[t]{0.47\linewidth}
			\centering
			\includegraphics[width=2.1cm]{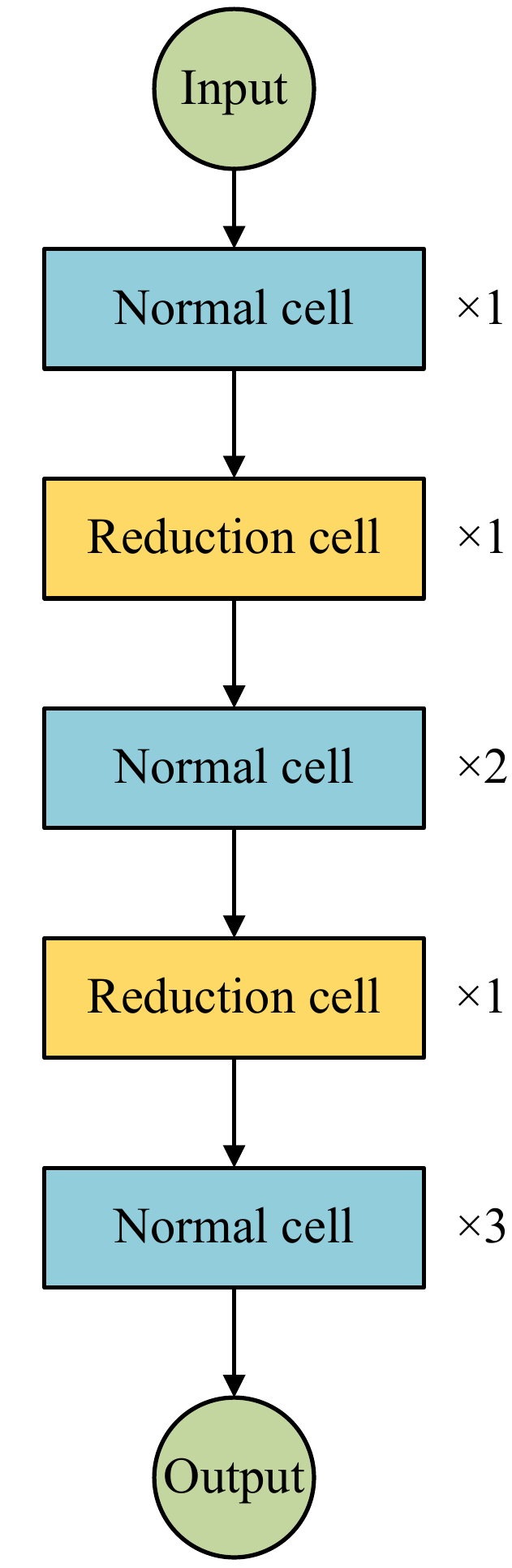}
			
		\end{minipage}
	}
	\centering
	\subfloat[ \ ]{
		\centering
		\begin{minipage}[t]{0.5\linewidth}
			\centering
			\includegraphics[width=2.1cm]{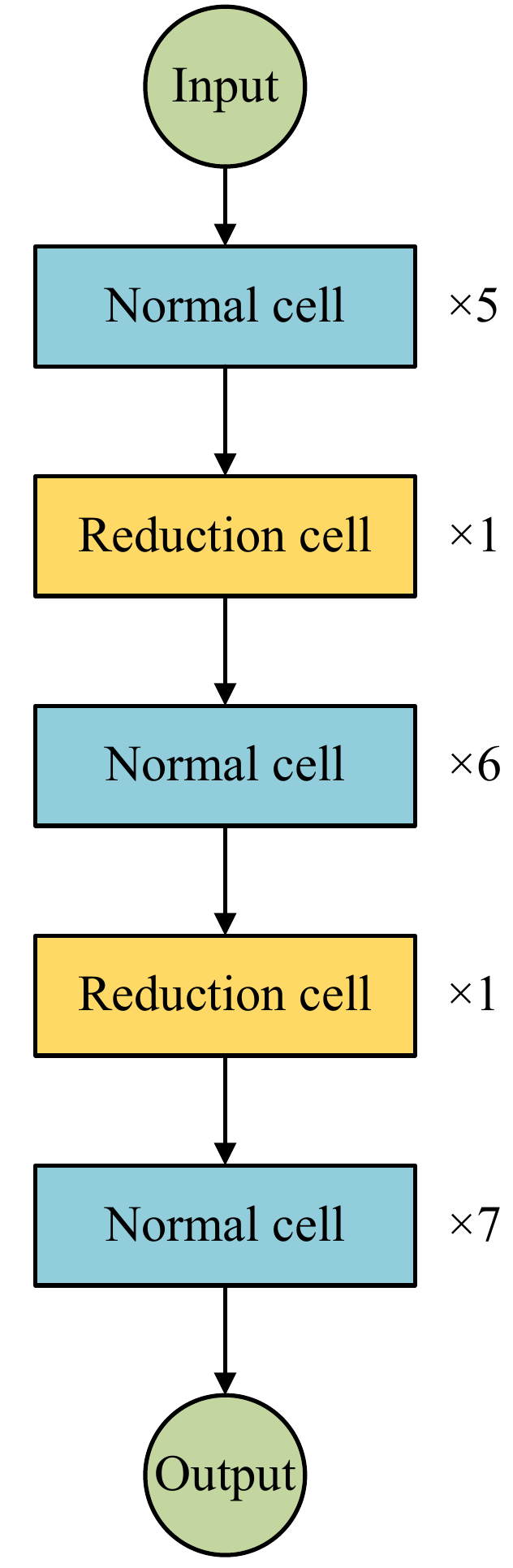}
		
		\end{minipage}
	}
	
	\caption{The network structure on the process of searching and training based on normal cell and reduction cell, (a) describes the searching network, (b) describes the evaluation network.}
	\label{f07}
\end{figure}
\subsection{Implementation Details}
\subsubsection{Network structure}
The network consists of the normal cells and the reduction cells. There are two input nodes, four intermediate nodes, and one output node in each cell. One-third and two-thirds of the network are reduction cells and the rest are normal cells. The operation space between nodes has eight candidate operations, which are $ 'none' $, $ 'maxpool\_3\times3' $, $ 'avgpool\_3\times3' $, $'skip\_connect'$, $'sepconv\_3\times3'$, $'dilconv\_3\times3'$, $'sepconv\_5\times5'$, $'dilconv\_5\times5'$.

\subsubsection{Experimental methods}
The search structure and evaluation structure are shown in Fig. \ref{f07} (a) and Fig. \ref{f07} (b). In the searching stage, eight cells are stacked into a network, and some epochs are run  on CIFAR-10 to search the cell structures. The searched normal cells and reduction cells are used to generate an evaluation structure by stacking again. Then, the network parameters were trained to test its performance.

\subsubsection{Parameter setting}
The epoch of network search was set to 80, batchsize was 96, cross entropy loss was used to calculate $\mathcal{L}_{train}$ and $\mathcal{L}_{val}$. The SGD optimizer was used to update network parameter $\omega$, the initial learning rate was 0.025, momentum was 0.9, and weight decay was 0.0003. The learning rate was gradually decreased to 0 by using cosine annealing strategy. The Adam optimizer was used to update the network architecture parameter $\alpha$ with an initial learning rate of 0.0006, a momentum of (0.5, 0.999), and a weight decay of 0.001.
\subsection{Experimental Results and Analysis}
\subsubsection{Analysis of searched cells}
\begin{figure*}[htbp]
	\centering	
	\subfloat[Normal cell]{
		\centering
		\begin{minipage}[t]{0.5\linewidth}
			\centering
			\includegraphics[width=8cm]{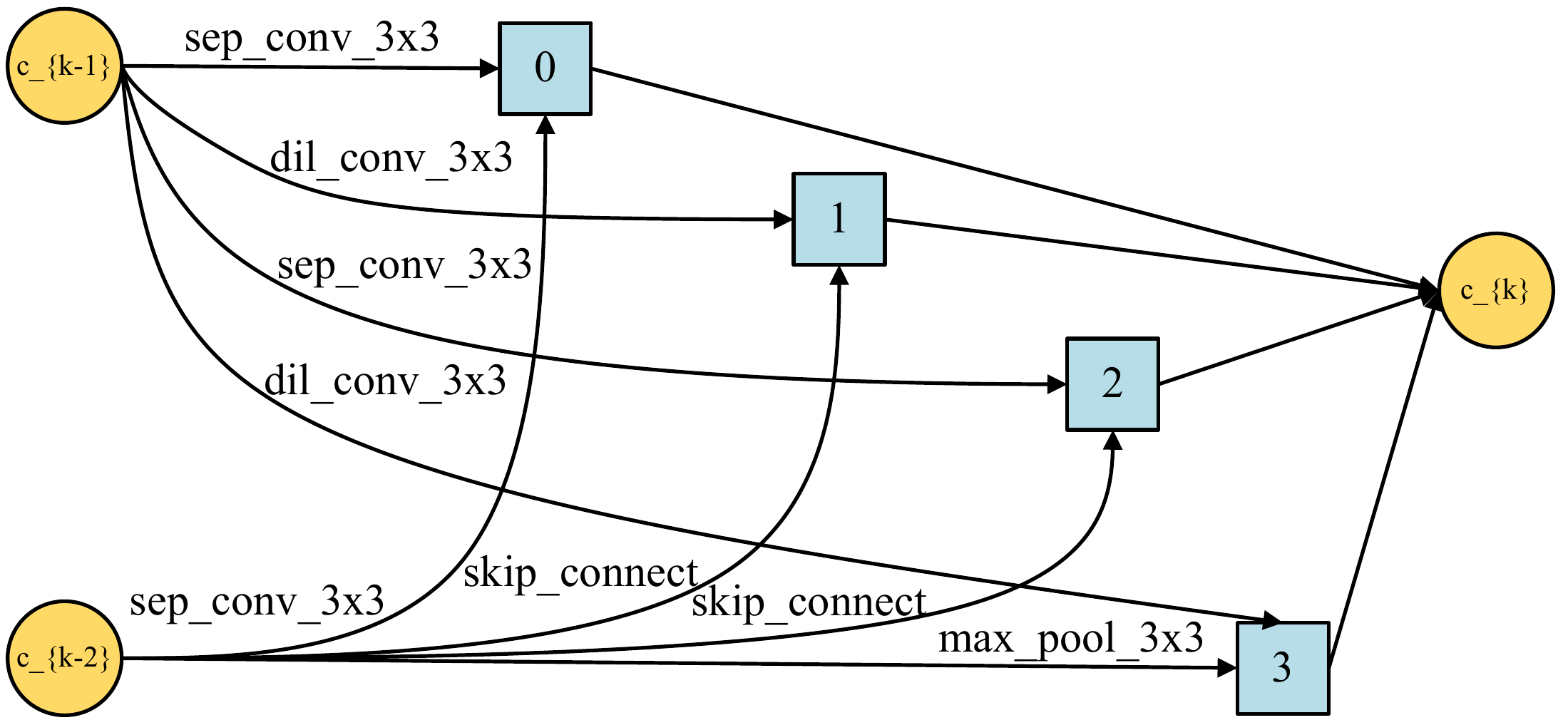}
		
		\end{minipage}
	}
	\centering
	\subfloat[Reduction cell]{
		\centering
		\begin{minipage}[t]{0.5\linewidth}
			\centering
			\includegraphics[width=8cm]{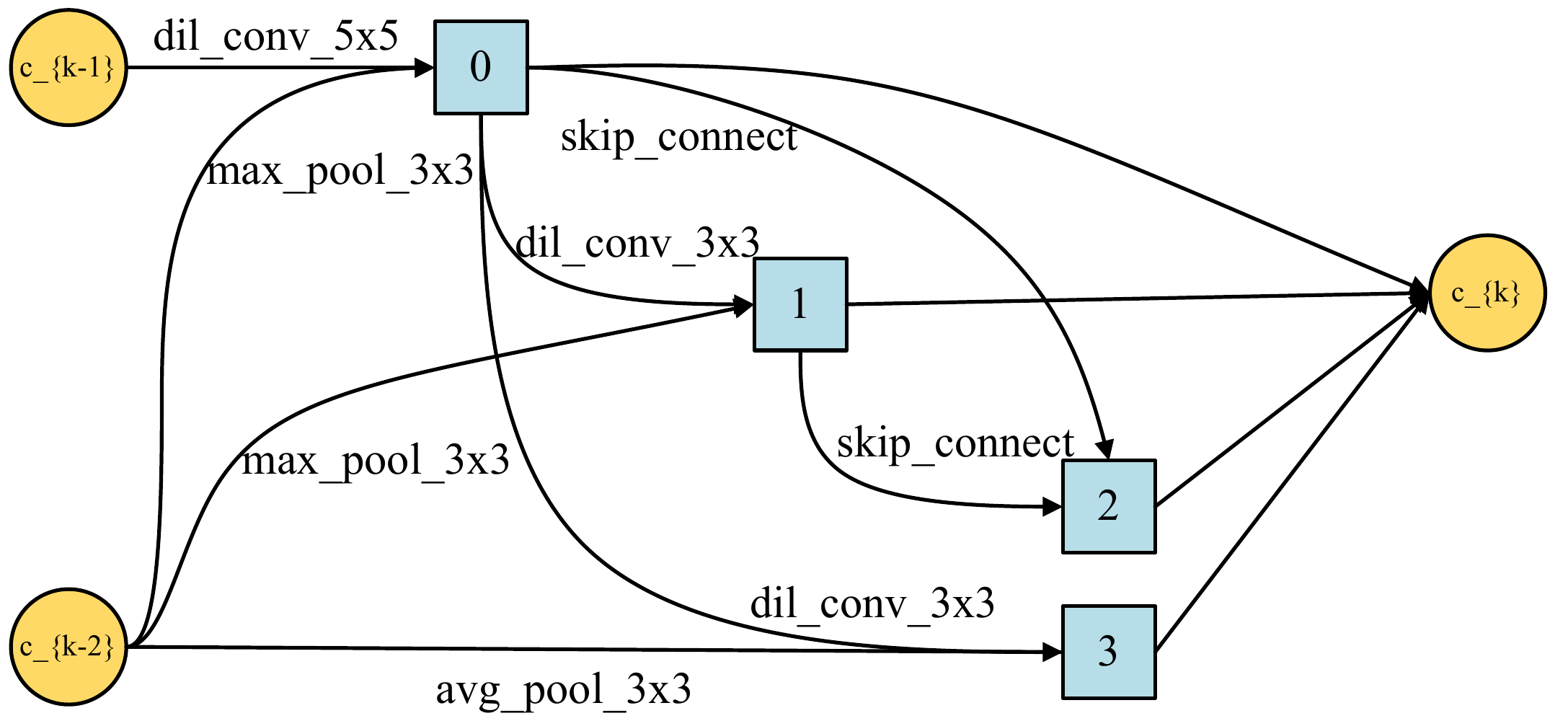}
		
		\end{minipage}
	}
	
	\caption{The structures of searched cells on CIFAR-10 by ADARTS.}
	\label{f08}
\end{figure*}

The searched cell structures by ADARTS are shown in Fig. \ref{f08}. Compared with final architecture in DARTS, the number of skip connections in the network architecture obtained by ADARTS is less and more stable, which can effectively alleviate the problems of gradient disappearance and network degradation. At the same time, more useful feature information can be extracted to enhance the performance of the network.

\subsubsection{Evaluation on CIFAR-10 and CIFAR-100}
Twenty cells are stacked into a new network for performance evaluation and the structure of the evaluation network is shown in Fig. \ref{f07} (b). The network was trained for 600 epochs with a batchsize of 64. SGD optimizer was used to update network parameters with a momentum of 0.9 and a weight decay of 0.0003. The initial learning rate was 0.017. In addition, the cosine annealing strategy was used to decrease the learning rate to 0. The cutout length of 16, auxiliary weight of 0.4, and drop path probability of 0.3 were applied to prevent overfitting.  

We compare the accuracy of ADARTS on CIFAR-10 and CIFAR-100 with some hand-designed networks, RL, evolution, and other gradient-based search methods. The results are given in Table \ref{Tab03}. We can see from this table that ADARTS achieved 2.46\% classification error on CIFAR-10 and 17.06\% classification error on CIFAR-100. More important, ADARTS only uses 0.2 GPU days, which is much faster than NASNet-A and AmoebaNet-B, and also faster than other gradient-based search methods. Compared with the baseline DARTS, ADARTS speeds up the search speed while also improving the classification accuracy by 0.3\% on CIFAR-10 and 0.48\% on CIFAR-100. Compared with GDAS, ADARTS achieves better classification accuracy within the same search time. Compared with PCDARTS, ADARTS can reduce the instability of selecting channels randomly so that it achieves better performance. Also, the parameters of ADARTS are 2.9 million which is less than most other algorithms. This indicates that ADARTS achieves better classification accuracy and faster searching speed.

Moreover, the network structure searched on CIFAR-10 is transferred to CIFAR-100 for testing and performs well, indicating that the searched network architecture have strong robustness.

\begin{table*}[]

	\centering
	
	\caption{Classification error rate of ADARTS on CIFAR-10 and CIFAR-100.}
	
	\label{Tab03}
	\scalebox{1}{
	\begin{tabular}{cccccccccc}
		
		\toprule
		
		\multirow{2}{*}{Architecture} & \multirow{2}{*}{Params(M)}& \multirow{2}{*}{Search Cost(GPU days)} & \multicolumn{2}{c}{Test Error} & \multirow{2}{*}{Search Method} \\
		
		\cmidrule(r){4-5}
		
		\ \ & & & CIFAR-10(\%)     &  CIFAR-100(\%)

		\\
		
		\midrule
		
		ResNet(depth=110) \cite{he2016deep}    & 1.7               & -    &6.43           & 25.16      & Manual        \\
		ResNet(depth=1202) \cite{he2016deep}      & 10.2           & -  &7.93          & 27.82          & Manual        \\
		DenseNet-BC \cite{huang2017densely}                & 25.6               & -    &3.46                  & 17.18     & Manual        \\
		VGG \cite{simonyan2014very}                       & 20.1               & -    &6.66            & 28.05       & Manual        \\
		MobileNetV2 \cite{sandler2018mobilenetv2}                 & 2.2               & -   &4.26            & 19.20       & Manual        \\
		
		\midrule
		Genetic CNN \cite{xie2017genetic}                  & -         & 17   &7.10           & 29.05        & evolution        \\
		AmoebaNet-B \cite{real2019regularized}            & 2.8         &3150   &2.55          & -       & evolution        \\
		
		Hireachical Evolution \cite{liu2018hierarchical}       &15.7      & 300       &3.75       & -         & evolution        \\	
		CARS \cite{yang2020cars}          &2.4                    & 0.4              & 3.00              &-         & evolution        \\
		\midrule	
		ENAS \cite{pham2018efficient}               & 4.6        &0.5      &2.89              & 19.43     & RL        \\
		NASNet-A \cite{zoph2018learning}               & 3.3        & 1800       &3.41             & -    & RL        \\
		NASNet-A + cutout \cite{zoph2018learning}      & 3.3       & 1800       &2.65              & -     & RL        \\
		SMASH \cite{brock2018smash}               & 16        & 1.5       &4.03              & -     & RL        \\
		\midrule
		DARTS (first order) + cutout \cite{liu2018darts}               & 3.3        & 1.5       &3.00              & 17.76     & gradient-based        \\
		DARTS (second order) + cutout \cite{liu2018darts}               & 3.3        & 4.0      &2.76              & 17.54     & gradient-based        \\
		GDAS \cite{dong2019searching}               & 3.4        & 0.2       &3.87              & 19.68     & gradient-based        \\
		GDAS+cutout \cite{dong2019searching}               & 3.4        & 0.2       &2.93        & 19.38     & gradient-based        \\
		SNAS + cutout \cite{xie2018snas}               & 2.8        & 1.5      &2.85        & 17.55     & gradient-based        \\
		P-DARTS + cutout \cite{chen2019progressive}               & 3.4        & 0.3    &2.50        & 17.20     & gradient-based        \\
		PCDARTS + cutout \cite{xu2019pc}               & 3.6       & 0.1    &2.57        & -     & gradient-based        \\
		
		FairDARTS + cutout \cite{chu2020fair}               & 2.8        & 0.4    &2.54        & 17.61     & gradient-based        \\
		SDARTS-ADV + cutout \cite{chen2020stabilizing}               & 3.3        & 1.3    &2.61        & 16.73     & gradient-based        \\
		EoiNAS + cutout \cite{zhou2021exploiting}               & 3.4        & 0.6    &2.50        & 17.30     & gradient-based        \\
		\midrule
		ADARTS               & 2.9        & 0.2    &3.70        & 18.21     & gradient-based        \\
		ADARTS + cutout               & 2.9        & 0.2    &2.46        & 17.06     & gradient-based        \\
		\bottomrule
		
	\end{tabular}
}
\end{table*}

\begin{figure*}[htbp]
	\centering	
	\subfloat[Skip connections in normal cell]{
		\centering
		\begin{minipage}[t]{0.45\linewidth}
			\centering
			\includegraphics[width=5.5cm]{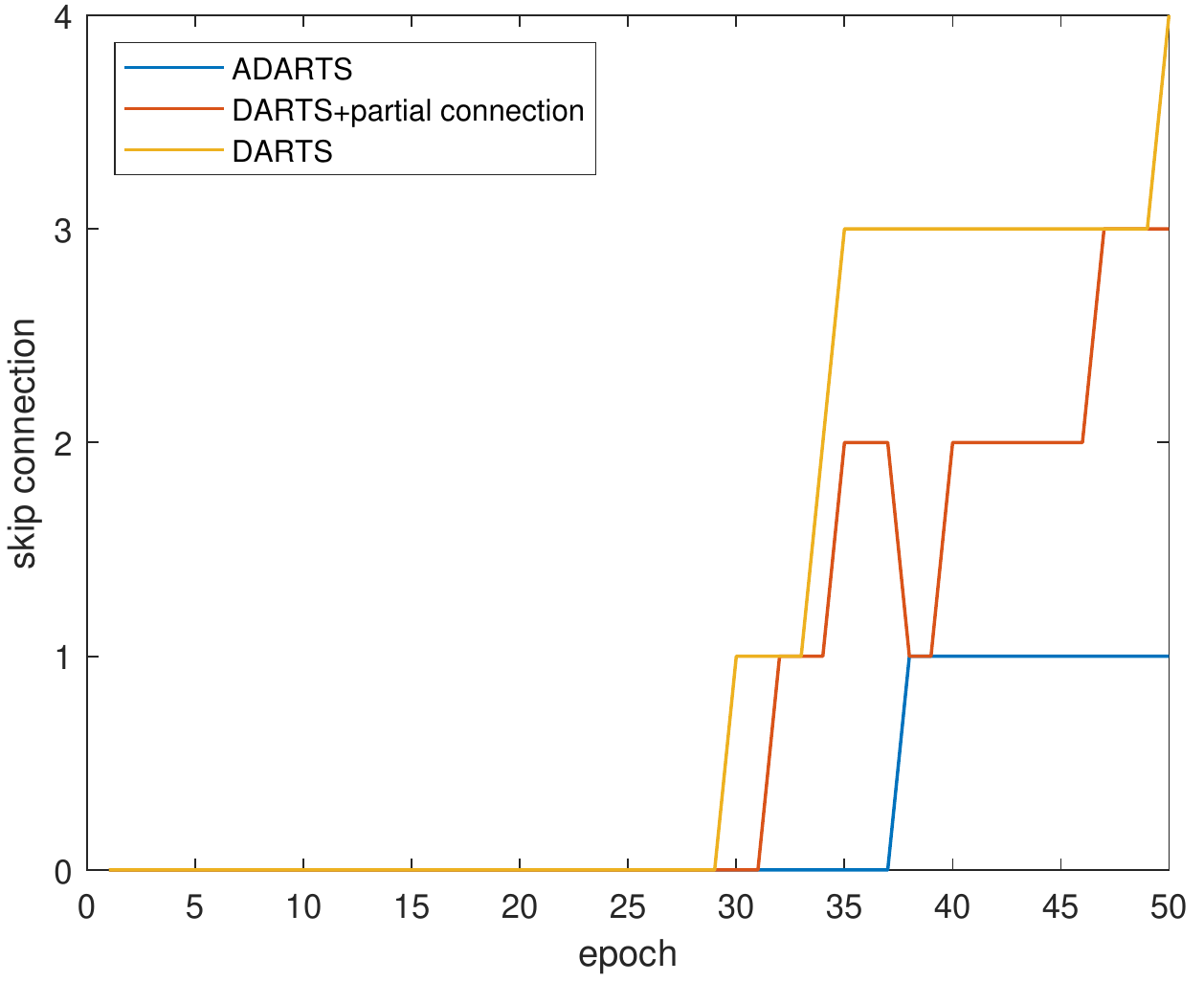}
	
		\end{minipage}
	}
	\centering
	\subfloat[Skip connections in reduction cell]{
		\centering
		\begin{minipage}[t]{0.45\linewidth}
			\centering
			\includegraphics[width=5.5cm]{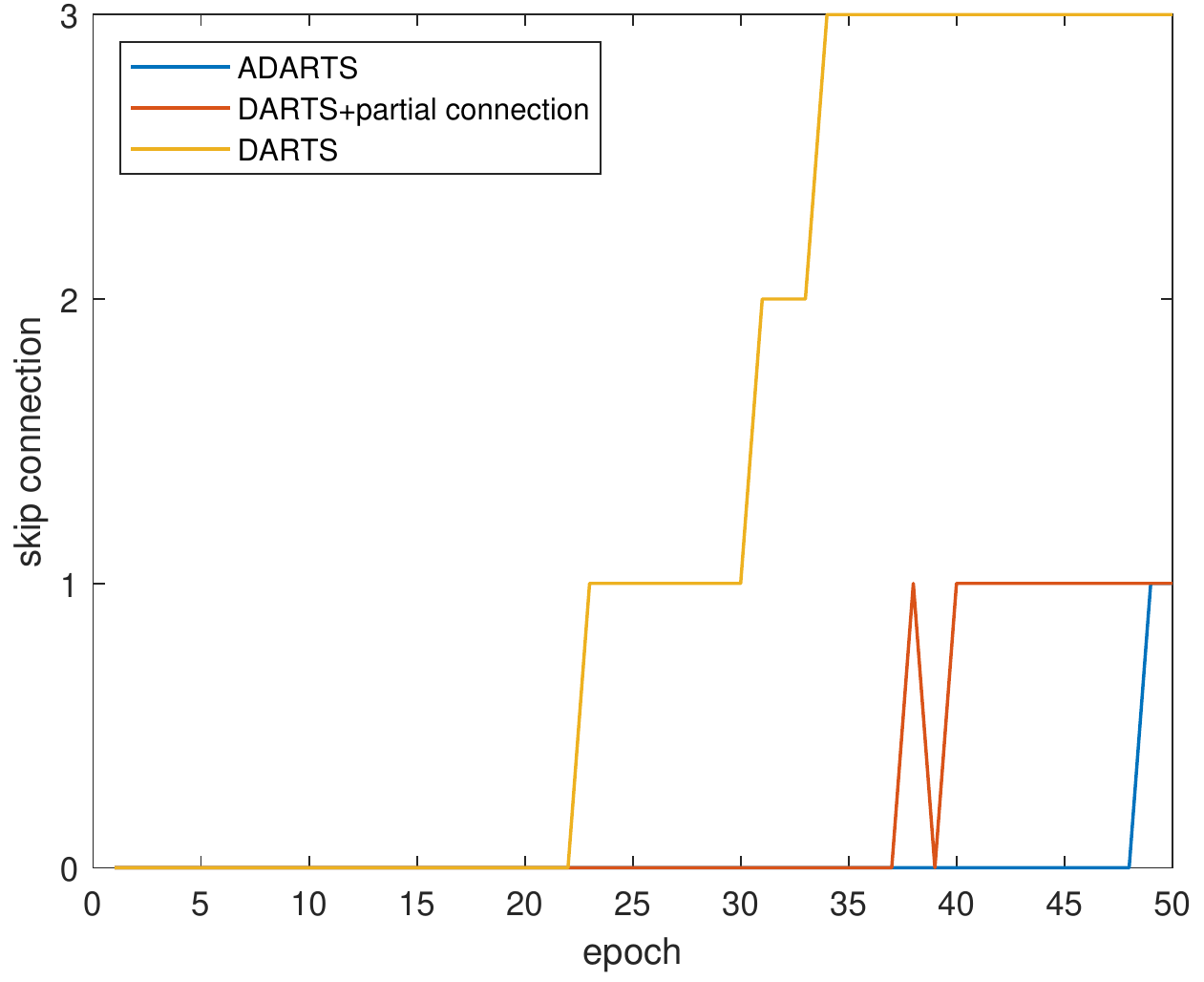}
			
		\end{minipage}
	}
	
	\caption{The number of skip connections of cells searched on CIFAR-10 by ADARTS.}
	\label{f09}
\end{figure*}
\subsection{Ablation Studies}
\subsubsection{Analysis of skip connection}
In DARTS, too many skip connections lead to low network performance, which is caused by unfair competition between operations. The parameters of the weight-equipped operations are not trained well in the early stage of the search and the weight-free operations accumulate too many advantages. There will be more and more skip connections in the network with the increasing of training epochs. The number of skip connections 
is shown in Fig. \ref{f09}. In this figure, we can see that four skip connections appeared when DARTS searched to 50 epochs, which greatly reduces the feature information extracted by the network, and results in low performance. When partial connection method is added to DARTS, the number of feature channels in the operation space is reduced, which can weaken the unfair competition in the early stage and reduce the accumulation of advantages of weight-free operations. It can also find that the number of skip connections of DARTS with partial connection is less than that of DARTS. However, the random channel selection also leads to the instability of search, so the channel attention mechanism can select more important channels and strengthen feature information to make the search process more stable. Only in the middle and late stages of the search, the network allows a few skip connections to alleviate the problem of gradient disappearance. All in all, ADARTS can effectively solve the problem of too many skip connections and obtain more stable and better network structures.
\subsubsection{Analysis of channel proportion 1/K}

\begin{figure}[h]
	\centering
	\includegraphics[width=7cm]{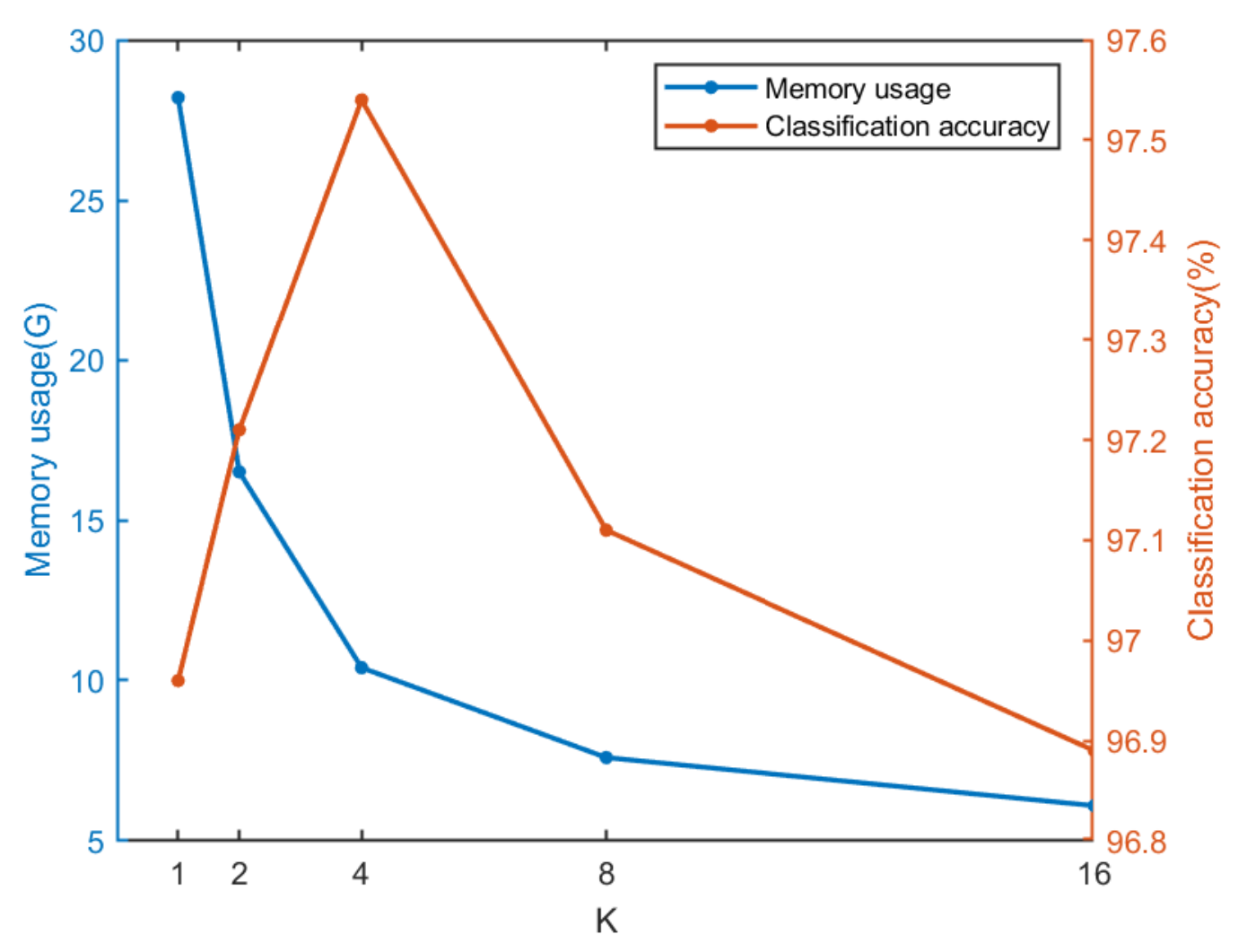}
	\caption{The memory usage and classification accuracy with different $K$ values. }
	\label{f10}
\end{figure}

To verify the influence of the $K$ value on memory usage and classification accuracy, and find the best $K$ value, we set different values of $K$=\{1,2,4,8,16\} for experimental analysis since the initial number of channels for input data is 16. Fig. \ref{f10} shows the experimental results with different $K$ values. With the increase of $K$, the memory usage decreases continuously, while the classification accuracy of the network improves first and then decreases. When $K$ is set to 4, the classification accuracy of the searched network structure is the highest, and the memory usage decreases greatly, which speeds up the search speed. This shows that partial channel connection based on attention mechanism can more effectively search network structure with better performance. However, the number of channels entering the search space should not be too small, otherwise the network can not get enough feature information during training, and thus can not get a good network structure.
\subsubsection{Analysis of methods of ADARTS}
To verify the effectiveness of attention mechanism and partial connection on architecture searching, we use DARTS with random partial connection, DARTS with attention mechanism and ADARTS to search on CIFAR-10, and test the searched network structures on CIFAR-10 and CIFAR-100. The experimental results are shown in Table~\ref{Tab04}. It is obvious that the search cost is greatly reduced through partial channel connection. In addition, the test classification errors also decrease on CIFAR-10/CIFAR-100 when using attention mechanism or partial channel connection. Moreover, ADARTS which combines attention mechanism and partial channel connection achieves the smallest parameters, fast search speed, and the best classification accuracy among all the methods. It shows that ADARTS can speed up the search process and improve the performance of the searched networks.

\begin{table}[!h]
	
	\centering
	
	\caption{The ablation studies on CIFAR-10 and CIFAR-100.}
	
	\label{Tab04}
	\scalebox{0.77}{
		\begin{tabular}{cccccccccc}
			\toprule
			\multirow{2}{*}{Method} & \multirow{2}{*}{Params}& \multirow{2}{*}{Search Cost} & \multicolumn{2}{c}{Test Error}  \\
			\cmidrule(r){4-5}
			\ \ &(M)&(GPU days) &  CIFAR-10    &  CIFAR-100
			\\
			
			\midrule
			
			DARTS    & 3.3               & 4.0    &2.76\%           & 17.54\%          \\
			DARTS + random partial connection     & 3.4           & 0.1  &2.71\%          & 17.42\%            \\
			DARTS + attention mechanism     & 3.1               & 0.6    &2.67\%           & 17.47\%          \\
			ADARTS    & 2.9               & 0.2    &2.46\%           & 17.06\%          \\
			\midrule
		\end{tabular}
	}
\end{table}

\subsubsection{Analysis of batchsize}
Because the increase of batchsize will lead to the decrease of the step of parameter adjustment in the mini-batch gradient descent, the learning rate should be appropriately increased \cite{goyal_accurate_2018,krizhevsky2014one}.
Thus, we set several groups of learning rate and batchsize to verify the effect of increasing batchsize appropriately. The experimental results are shown in Table \ref{Tab02}. When the learning rate is set to 0.01, the accuracy of ADARTS increases as batchsize increases from 8 to 16. Besides, when learning rate is set to 0.025, the accuracy of ADARTS increases as batchsize increases from 32 to 96. Moreover, too small batchsize leads to a significant decrease in stability and accuracy of ADARTS. Therefore, increasing batchsize appropriately can improve the performance of ADARTS.
\begin{table}[!h]
	
	\centering
	
	\caption{The classification accuracy of ADARTS on various learning rate and batchsize.}
	
	\label{Tab02}
	\scalebox{0.8}{
		\begin{tabular}{cccccccccc}
			\toprule
			\multirow{2}{*}{Method} 
			& \multicolumn{2}{c}{Learning rate=0.01} 	& \multicolumn{3}{c}{Learning rate=0.025}  \\
			\cmidrule(r){2-3}\cmidrule(r){4-6}
			
			\ \ &batchsize=8 &batchsize=16    &  batchsize=32  &batchsize=64 &batchsize=96
			\\
			
			\midrule
			
			\\
			ADARTS    & 96.89\%               & 97.11\%   &97.36\%  &97.51\%          & 97.54\%           \\
			\midrule
		\end{tabular}
	}
\end{table}

\section{Conclusion}
In this paper, ADARTS is proposed to solve the performance crash due to too many weight-free operations in late search and instability of the search structure caused by low memory utilization. The channel attention and partial channel connection are used to solve problems. The channels with higher weights in the channel attention weight are sent to the operation space. These methods can reduce the consumption of calculation in the operation space to speed up the search speed and improve the stability of searching network structure. Besides, the number of weight-free operations in structure can be controlled so that the searched network can perform well. The algorithm is applied to CIFAR-10 and CIFAR-100 for image classification and the experiment shows that ADARTS can search out the network architecture quickly and stably.
In the future work, we will further enhance the stability of channel selection and reduce redundant feature information. We are considering introducing some typical feature selection techniques to feature map selection problems in deep neural networks. In addition, we will conduct the proposed method on some industrial datasets for network search, and use the obtained network structure in facial recognition, target detection, semantic recognition, etc.

\bibliographystyle{IEEEtran}

\end{document}